\documentclass[runningheads]{llncs}

\usepackage[T1]{fontenc}
\usepackage{cite}
\usepackage{amsmath,amssymb,amsfonts}
\usepackage{algorithmic}
\usepackage{graphicx}
\usepackage{textcomp}
\usepackage{xcolor}
\usepackage{svg}
\usepackage{url} 
\usepackage{hyperref}
\usepackage{caption}
\usepackage{fancyhdr}
\usepackage{makecell}
\usepackage{subfigure}
\usepackage{blindtext}
\usepackage{booktabs}
\usepackage{multirow}
\usepackage{algorithm}
\usepackage{algorithmic}
\usepackage{diagbox}
\usepackage{graphicx}

\begin{document}
\title{Tailored Federated Learning: Leveraging Direction Regulation \& Knowledge Distillation}

\author{Huidong Tang\inst{1,3} \and
Chen Li*\inst{2} \and
Huachong Yu\inst{3} \and
Sayaka Kamei\inst{3} \and
Yasuhiko Morimoto\inst{3}}

\authorrunning{H. Tang et al.}
\institute{Shandong Xiehe University, Jinan, China\\
\email{tanghd24@163.com}\\ \and
Graduate School of Informatics, Nagoya University, Nagoya, Japan\\
\email{li.chen.z2@a.mail.nagoya-u.ac.jp}\\ \and
Graduate School of Advanced Science and Engineering, Hiroshima University, Higashi-Hiroshima, Japan\\
\email{huachongyu,s10kamei,morimo@hiroshima-u.ac.jp}}

\maketitle   

\begin{abstract}
Federated learning (FL) has emerged as a transformative training paradigm, particularly invaluable in privacy-sensitive domains like healthcare. However, client heterogeneity in data, computing power, and tasks poses a significant challenge. To address such a challenge, we propose an FL optimization algorithm that integrates model delta regularization, personalized models, federated knowledge distillation, and mix-pooling. Model delta regularization optimizes model updates centrally on the server, efficiently updating clients with minimal communication costs. Personalized models and federated knowledge distillation strategies are employed to tackle task heterogeneity effectively. Additionally, mix-pooling is introduced to accommodate variations in the sensitivity of readout operations. Experimental results demonstrate the remarkable accuracy and rapid convergence achieved by model delta regularization. Additionally, the federated knowledge distillation algorithm notably improves FL performance, especially in scenarios with diverse data. Moreover, mix-pooling readout operations provide tangible benefits for clients, showing the effectiveness of our proposed methods.

\end{abstract}

\section{Introduction}
Machine learning (ML) has become a pervasive force across diverse domains such as natural language processing, computer vision, and healthcare, owing much of its success to its capacity for learning from vast datasets \cite{li2023spotgan,li2024tengan}. However, traditional centralized learning approaches encounter significant hurdles, notably in navigating privacy concerns and complying with data protection regulations such as GDPR \cite{Murphy2018TheGD}. These challenges are particularly acute in sensitive sectors like healthcare, where safeguarding the privacy of medical data is paramount, yet ML advancements are imperative.

In response to these privacy concerns, Federated learning (FL) has emerged as a promising solution. FL enables multiple clients to collaboratively train a shared model using their decentralized local data \cite{McMahan2016CommunicationEfficientLO, McMahan2016FederatedLO, duan2020learning, huang2019patient, li2019distributed}. However, the heterogeneity of data, tasks, and computing power presents a significant bottleneck to FL performance \cite{Kairouz2019AdvancesAO, Li2019FederatedLC, Gao2022ASO}. Data heterogeneity, particularly when training data is non-IID (independent and identically distributed) across clients, can impede convergence and diminish performance compared to traditional centralized methods \cite{Zhao2018FederatedLW, Hsu2019MeasuringTE}, a challenge exacerbated with a growing number of clients. To address this, various approaches have been explored, including parameter regularization-based methods (e.g., FedProx \cite{Sahu2018FederatedOI}, FedPD \cite{Zhang2020FedPDAF}, FedDANE \cite{Li2019FedDANEAF}) and model momentum-based methods (e.g., SCAFFOLD \cite{Karimireddy2019SCAFFOLDSC}). Additionally, personalized federated learning (pFL) techniques such as FedBN \cite{Li2021FedBNFL}, LG-FedAvg \cite{Liang2020ThinkLA}, CD2-pFed \cite{Shen2022CD2pFedCD}, and Ditto \cite{Li2020DittoFA} have been explored. However, these efforts have not consistently achieved optimal performance with low communication costs for bandwidth-limited edge devices.

To address these challenges, we propose a federated optimization algorithm that integrates model delta regularization, personalized models, federated knowledge distillation, and mix-pooling. Our model delta regularization computes the optimization direction of the shared global model weights on the server and introduces an $L_2$ regularization term to align local model updates with the global model, significantly reducing communication costs compared to existing momentum-based methods. We leverage personalized models and federated knowledge distillation to address extreme heterogeneity settings, where both data and tasks vary significantly among clients. In such scenarios, personalized models do not participate in aggregation, and the global model transfers knowledge between personalized models via loss. Additionally, we introduce a mix-pooling layer to accommodate different client preferences. The main contributions are summarized as follows:
\begin{itemize}
\item{\bf Performance with low communication costs:} Our model delta regularization method ensures FL performance is maintained with minimal communication costs, making it suitable for bandwidth-limited edge devices.
\item{\bf Suitable for extreme settings:} The combination of personalized models and federated knowledge distillation enhances overall performance, particularly in extreme settings.
\item{\bf Catering to different client preferences:} The mix-pooling method accommodates diverse client preferences, benefiting specific clients while supporting federated distillation.
\end{itemize}

\section{Related Works}
\subsection{Challenges and Solutions in FL Algorithms}
FedAvg \cite{McMahan2016CommunicationEfficientLO} stands as one of the most renowned FL algorithms, addressing the critical issues of communication efficiency and client scalability. By implementing local aggregation among clients, it effectively reduces communication costs. However, it falls short in tackling the challenge of dataset heterogeneity, particularly non-IID datasets, which can significantly degrade global model performance due to divergent local model updates. To mitigate this, various solutions have been proposed. FedProx \cite{Sahu2018FederatedOI} combines a proximal term and local update iteration to tightly constrain local updates around the global model. FedOpt \cite{Reddi2020AdaptiveFO} employs a recursive training method with momentum and adaptive learning to expedite model convergence. SCAFFOLD \cite{Karimireddy2019SCAFFOLDSC} introduces a correction term to the local model update formula, based on the direction of the model update. FedDyn \cite{Acar2021FederatedLB} guides client updates along the global model update path, albeit with additional control variables. Despite their contributions, few of these methods manage to uphold performance standards while keeping communication costs low.

\subsection{Personalized Approaches for Handling Non-IID Datasets}
Another avenue for addressing non-IID datasets is pFL, where each client updates using heterogeneous weights derived from global model fine-tuning or knowledge distillation. Such approaches not only ensure the robustness of the global model but also align with the local data distribution. For instance, FedBN \cite{Li2021FedBNFL} specializes in handling feature shift non-IID problems with fast convergence, leveraging a batch normalization layer within local models to avoid communication and aggregation via the global model. FedRep \cite{Collins2021ExploitingSR} utilizes the shared representation of non-IID data, enabling clients to jointly learn this representation before refining their unique heads. FedMD \cite{Li2019FedMDHF} employs knowledge distillation as a basis for FL methods across different models. FedGKD \cite{Yao2021LocalGlobalKD} employs client-side knowledge distillation instead of relying on public datasets for FL guidance. However, these approaches often necessitate additional tuning or knowledge distillation, which poses challenges for bandwidth-limited edge devices. 

To address these issues, we introduce a comprehensive approach combining model delta regularization, personalized models, federated knowledge distillation, and mix-pooling to uphold performance while minimizing communication costs. Our method demonstrates adaptability in extreme heterogeneous settings.

\section{Preliminaries}
\subsection{Non-IID Data}
IID data assumes that all data within the same dataset are sampled from the same underlying data distribution, and that each data point is independent of all others as follows:
\begin{align}
&\forall x^{(i)} \sim \mathcal{D},~~\forall i \neq j,~~\text{and}~~p\left(x^{(i)}, x^{(j)}\right)=p\left(x^{(i)}\right) p\left(x^{(j)}\right),
\end{align}
where $\mathcal{D}$ represents the independently distributed dataset. Non-IID data refers to data that does not satisfy any of the independent or identical distribution. In the FL scenarios, based on the relationship between the input distribution $P(X)$ and the label distribution $P(Y \mid X)$ across clients, non-IID data can be categorized into feature distribution skew, label distribution skew, same label, different features, and same features, different label. Formally, they can be expressed as:
\begin{align}
&P_i(X) \neq P_j(X), P_i(Y \mid X) = P_j(Y \mid X),\\
&P_i(Y) \neq P_j(Y), P_i(X \mid Y) = P_j(X \mid Y),\\
&P_i(Y) = P_j(Y), P_i(X \mid Y) \neq P_j(X \mid Y),\\
&P_i(X) = P_j(X), P_i(Y \mid X) \neq P_j(Y \mid X). 
\end{align}
Non-IID data can introduce model bias, leading to poor model performance on unseen data and slow convergence. Training may converge slowly or fail to converge altogether due to discrepancies across clients. Consequently, non-IID data presents a significant challenge in FL settings.

\subsection{FedAvg Algorithm}
FedAvg \cite{McMahan2016CommunicationEfficientLO} operates through iterative training of the global model. Initially, the global model is transmitted from the server to a subset of clients, randomly chosen in each communication round. These clients update their respective models based on their local data. Subsequently, the updated models from each client are sent back to the server, where they are averaged to produce a revised global model. The objective function can be described as follows:
\begin{align}
\min_{\mathbf{w} \in \mathbb{R}^d} F(\mathbf{w}) = \sum_{k=1}^{K} \frac{n_k}{n}F_k(\mathbf{w}), 
\end{align}
where $\mathbf{w}$ represents the model parameters, $F$ denotes the global loss function, $F_k$ stands for the local loss function of the $k$-th client, $K$ signifies the total number of clients, $n_k$ represents the number of samples on the $k$-th client for training, and $n$ indicates the total number of training samples available.

\subsection{Knowledge Distillation}
In knowledge distillation (KD), a learning method where a student model learns from a teacher model \cite{Bucila2006ModelC, Hinton2015DistillingTK}, the loss function is as follows:
\begin{align}
&\mathcal{L}=(1-\lambda) \mathcal{L}_{C E}\left(q^S, y\right)+\lambda \mathcal{L}_{K L}\left(q_\tau^S, q_\tau^T\right),\\
&q_\tau(i)=\frac{\exp \left(z_i / \tau\right)}{\sum_j \exp \left(z_j / \tau\right)},
\end{align}
where $\mathcal{L}_{CE}$ represents the cross-entropy loss, quantifying the disparity between the student predictions and the true labels. $\mathcal{L}_{KL}$ denotes the Kullback-Leibler (KL) loss, measuring the discrepancy between the soft labels of the student and teacher models. $\lambda$ serves as a weighting factor. Notably, $q_\tau^S$ and $q_\tau^T$ correspond to the softmax outputs of the student and teacher models, respectively, with a temperature parameter $\tau$, while $z_i$ represents the $i$-th logits vector value.

\section{Tailored Federated Learning}
\subsection{Model Delta Regularization}
A direct approach to tackle the issue of heterogeneity involves aligning the update directions of local models across all clients. To achieve this, we calculate the weight difference of the global model, denoted as ${\delta}^t$, between each update round. Subsequently, this computed ${\delta}^t$ is transmitted to the clients alongside the global model parameters $w^t$ in round ${t+1}$. Following the FedAvg method \cite{McMahan2016CommunicationEfficientLO}, we update the global model by computing the average of updates from selected clients. However, the limited number of clients involved in each training round fails to fully capture the overall variance in weight updates across all clients. To address this limitation, we augment our approach by integrating historical deltas n the following manner:
\begin{align}
\boldsymbol {\delta}^t \leftarrow \begin{cases}\text { Option I. } & {\sum_{i=0}^{n} a_i \cdot \delta^{t-i}}, \\ \text { Option II. } & (1 - a) \cdot \delta^{t-1} + a \cdot \delta^t .\end{cases}
\end{align}
Here, Option I calculates the weighted average of $\delta$ over a restricted number of rounds, denoted by $n \leq 3$, with $\sum_{i=0}^{n} a_i = 1$, whereas Option II computes the moving average of $\delta$ with a decaying factor $0 < a < 1$ as training advances. To constrain the local model update directions towards the global $\delta$, we integrated an L2 regularization term. The optimization objective function is succinctly summarized as follows:
\begin{align}
\min_{w_k} \frac{1}{N} \sum_{i=1}^{N} Loss(w_k, x_{ki}, y_{ki}) + \frac{\mu}{2}\left|\left|f(w_k - w^t - \delta^t)\right|\right|^2,
\end{align}
\begin{align}
f(w_k - w^t - \delta^t) = \begin{cases} 0, \text{ if } (w_k - w^t) \cdot \delta^t >= 0, \\ {w_k - w^t - \delta^t}, \text{ otherwise}, \end{cases}
\end{align}
where $N$ represents the total number of rounds of local training on client $k$, $\mu$ serves as a scaling parameter controlling the L2 regularization term, and the function $f$ acts as a filter. This filter ensures that the L2 regularization term solely constrains the local model weights updated opposite to the global $\delta$. The algorithm flow is outlined in Algorithm \ref{alg:Fedr}.

\begin{algorithm}
\caption{Procedure of the federated model delta regularization} 
\label{alg:Fedr} 
\begin{algorithmic}
\STATE Server executes:
\STATE Initialize model weight $w_0$, model delta $\delta_0$
\FOR {each round $t$ = 1, 2...}
\STATE $\boldsymbol {\delta}^t \leftarrow \begin{cases}\text { Option I. } & {\sum_{i=0}^{n} a_i \cdot \delta^{t-i}},  \\ \text { Option II. } & (1 - a) \cdot \delta^{t-1} + a \cdot \delta^t,\end{cases}$
\STATE $S_t \gets$ (all clients)
\FOR {each client $k \in S_t$ in parallel}
\STATE $w_k^{t+1} \gets$ ClientUpdate($k, w^t, \delta^t$)
\ENDFOR
\STATE $w^{t+1} \gets \sum_{k=1}^{K}\frac{n_k}{n}w_k^{t+1}$,
\ENDFOR
\newline
\STATE ClientUpdate($k, w$):
\FOR {for each epoch $i$ from $1$ to $E$ on client $k$}
\STATE Loss $\gets \text{criterion}(w_k, x_k, y_k)$
\STATE Regularization loss $\gets \frac{\mu}{2}\left|\left|f(w_k - w^t - \delta^t)\right|\right|^2$
\STATE $w_k \gets \eta($Loss $+$ Regularization loss$)$
\ENDFOR
\end{algorithmic} 
\end{algorithm}

\subsection{Federate Knowledge Distillation} 
\begin{figure}[ht]
\centerline{\includegraphics[width=0.8\textwidth]{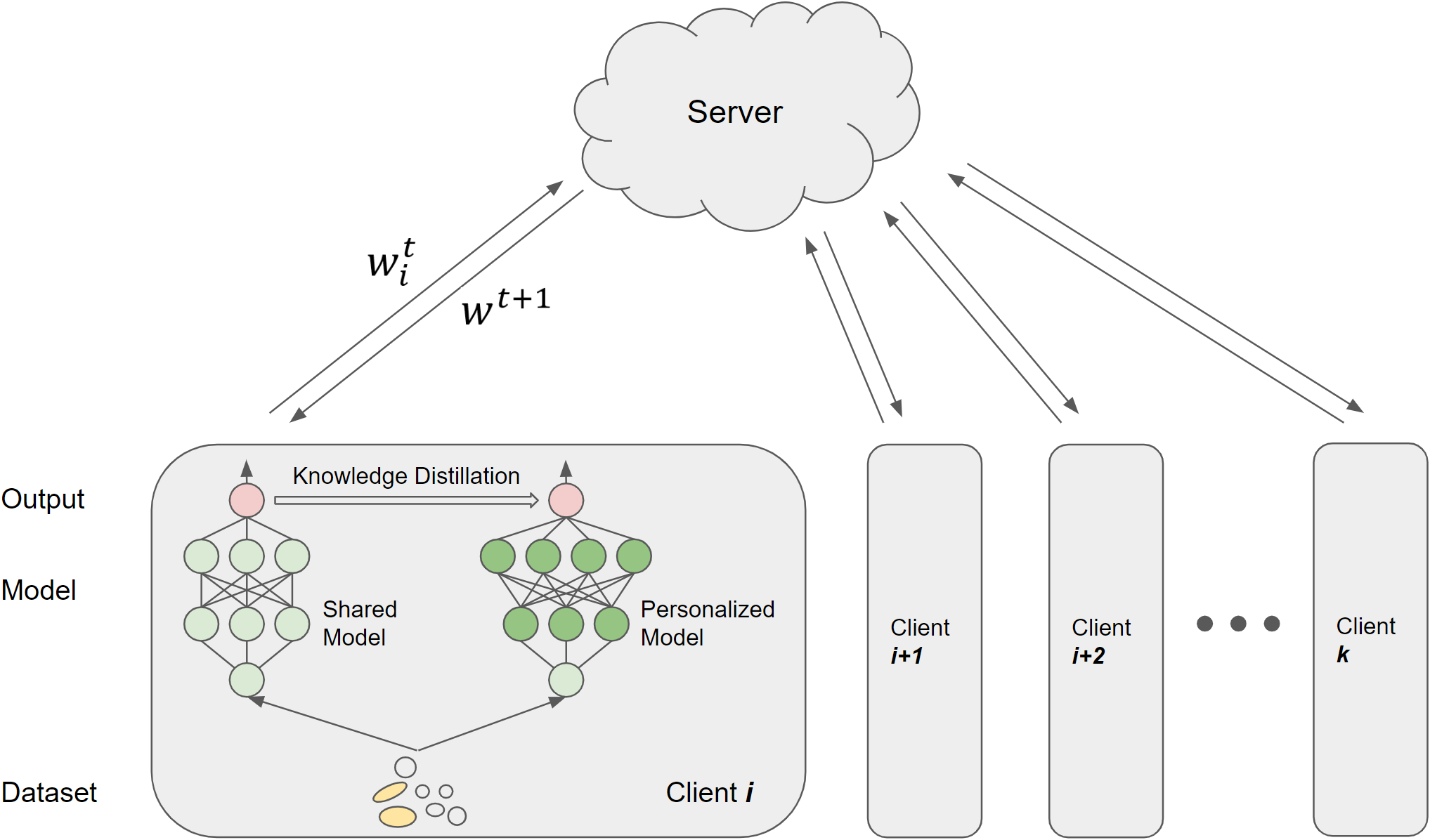}}
\caption{Overview of federate knowledge distillation.}
\label{fig:FedKD}
\end{figure}

To address highly heterogeneous tasks, where a single shared model is insufficient to accommodate the diverse data distributions and client tasks, we incorporate KD into federated learning. Specifically, we introduce a new model, denoted as $p$, for each client. These individual models undergo training alongside the global model through KD. However, they do not participate in the aggregation process at the server.

During each local update round, the loss function of each local model $p$ combines both the Cross-Entropy loss and the KL divergence loss \cite{li2024gxvaes}. This combined loss can be described as follows:
\begin{align}
&\mathcal{L}_{CE} = \text{criterion}(pred,\ y),
\end{align}
\begin{align}
&\mathcal{L}_{p} = (1 - \alpha) * \mathcal{L}_{CE} + \alpha * \mathcal{L}_{KL}(\text{softmax}(\frac{pred_p}{T}),\ \text{softmax}(\frac{pred}{T})).
\end{align}
Here, $\mathcal{L}_{CE}$ denotes the Cross-Entropy loss, and $\mathcal{L}_{KL}$ represents the KL loss. $y$ signifies the ground truth labels, while $pred$ and $pred_p$ denote the predictions of the global model and model $p$, respectively. $\alpha$ serves as the balancing factor, and $T$ is used to control the difficulty of the soft labels.

Our KD-based method addresses two main issues. Firstly, the need for re-training due to the replacement of the shared model on each client after aggregation is eliminated with our personalized models, reducing training costs. Secondly, the application of KD allows for the teacher model and the student model to differ, making it applicable to a wider range of real FL scenarios. Figure \ref{fig:FedKD} provides an overview of federated knowledge distillation.

\subsection{Mix-Pooling Readout} 
\begin{figure}[t]
\centerline{\includegraphics[width=0.6\textwidth]{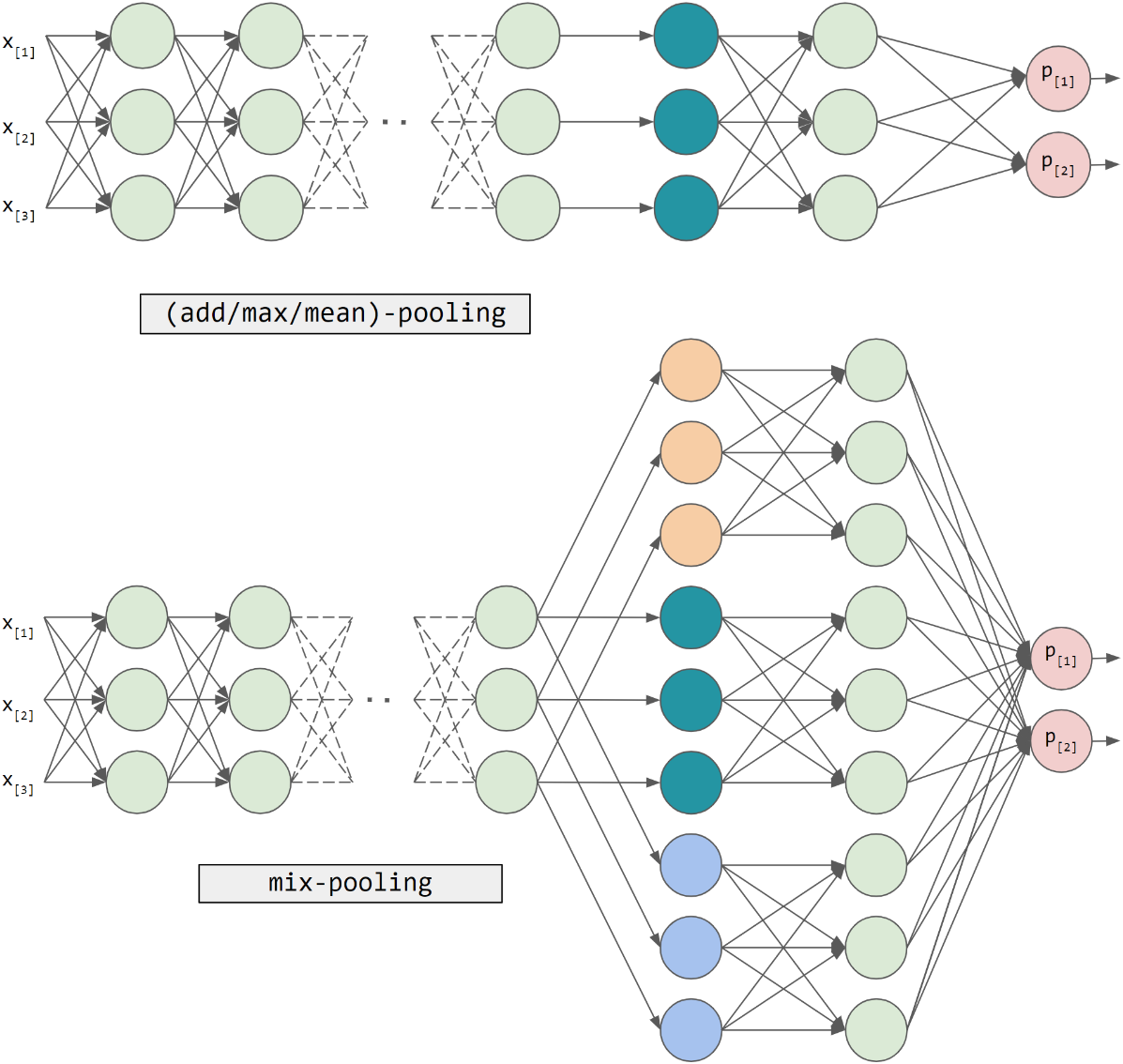}}
\caption{Overview of the mix-pooling layer structure.}
\label{fig:mix-pooling}
\end{figure}

For graph-based datasets, the readout operation is an important factor in graph neural networks. Ideally, the readout operation should be designed depending on the specific task and dataset. In FL settings, each client prefers different readout operations, which requires the above-mentioned specific readout operation design for each client. Therefore, we mix multiple pooling methods ("sum", "mean", and "max") for a hybrid readout layer, called \textit{mix-pooling}. 

Figure \ref{fig:mix-pooling} illustrates the concept of mix-pooling. The "sum" pooling method calculates the readout representation by summing over the representations of all nodes in the graph. This approach ensures that each node's contribution is equally weighted, resulting in a comprehensive summary of the graph's information. The "mean" pooling method computes the average of node representations, providing a representation that reflects the overall characteristics of the graph in a more balanced manner. The "max" pooling method selects the maximum value from the representations of all nodes in the graph, emphasizing the most prominent features present in the graph. By focusing solely on the highest value of each dimension across all node representations, "max" pooling captures the most salient aspects of the graph structure. Each pooling method offers a distinct approach to aggregating information from node representations, tailored to different aspects of the graph's structure and task requirements.

\section{Experiments}
\subsection{Datasets}
\begin{figure}[ht]
\centerline{\includegraphics[width=0.6\textwidth]{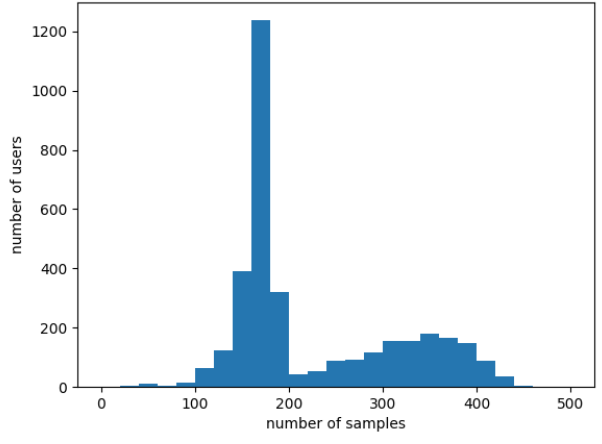}}
\caption{FEMNIST Dataset Details}
\label{fig:data_FEMNIST}
\end{figure}

We conducted experiments using three datasets: FEMNIST \cite{caldas2018leaf} and CIFAR-10 \cite{krizhevsky2009learning} for image classification tasks, and CIKM22Cup \cite{CIKM22} for graph-level blending tasks. Due to limited computational resources, we did not perform experiments on high-resolution medical images. The FEMNIST dataset \cite{caldas2018leaf} and CIFAR-10 \cite{krizhevsky2009learning} are well-known benchmarks for image classification. FEMNIST comprises 805,263 samples divided into 3,500 non-IID clients, as detailed in Figure \ref{fig:data_FEMNIST}. CIFAR-10 consists of 60,000 samples across 10 categories. We constructed non-IID distributions among clients using latent Dirichlet allocation (LDA). The CIKM22Cup \cite{CIKM22} dataset is specifically designed for heterogeneous tasks and comprises 13 clients. Table \ref{tab:cikm22cup_dataset_details} contains detailed information about the dataset.

\begin{table}[htbp]
\caption{CIKM22Cup Dataset Details}
\label{tab:cikm22cup_dataset_details}
\renewcommand\tabcolsep{15pt} 
\begin{center}
\begin{tabular}{ccclc}
\hline
\textbf{Client ID}&\textbf{Task type}&\textbf{Metric}&\textbf{Size}&\textbf{$b_i$}\\
\hline
1 & cls & Error rate & 1249 & 0.263789\\
2 & cls & Error rate & 181 & 0.289617\\
3 & cls & Error rate & 2219 & 0.355404\\
4 & cls & Error rate & 101 & 0.176471\\
5 & cls & Error rate & 188 & 0.396825\\
6 & cls & Error rate & 1101 & 0.261580\\
7 & cls & Error rate & 2228 & 0.302378\\
8 & cls & Error rate & 777 & 0.211538\\
9 & reg & MSE & 134706 & 0.059199\\
10 & reg & MSE & 109392 & 0.007083\\
11 & reg & MSE & 2268 & 0.734011\\
12 & reg & MSE & 608 & 1.361326\\
13 & reg & MSE & 70648 & 0.004389\\
\hline
\multicolumn{4}{l}{$b_i$ is the "isolated training" baseline.}
\end{tabular}
\end{center}
\end{table}

\subsection{Experimental Setup}
We employed the FederatedScope framework \cite{Xie2022FederatedScopeAC} for conducting our evaluation experiments. The model delta regularization method was evaluated across all three datasets. For the image datasets, we utilized the convnet2 architecture with a hidden layer width set to 2048. We created 200 clients, with 20\% of them participating in each round. Each client underwent local training for one epoch, and their updates were aggregated on the server. The learning rate was set to 0.1. Evaluation was performed every ten rounds, monitoring accuracy and loss across all clients. The total number of rounds conducted was 300. Regarding the graph dataset, we employed the GIN model with a layer width of 64 and utilized the "mean" pooling readout operation. Thirteen clients were utilized, with all of them participating in each round. Each client underwent local training for one epoch before their updates were aggregated on the server. Evaluation occurred every five rounds, assessing model improvement rate and loss across all clients. A total of 100 rounds were conducted.

\subsection{Experimental Results}
\begin{figure}[htbp]
\centering
\subfigure[FEMNIST test\_acc]{
\includegraphics[width=0.48\textwidth]{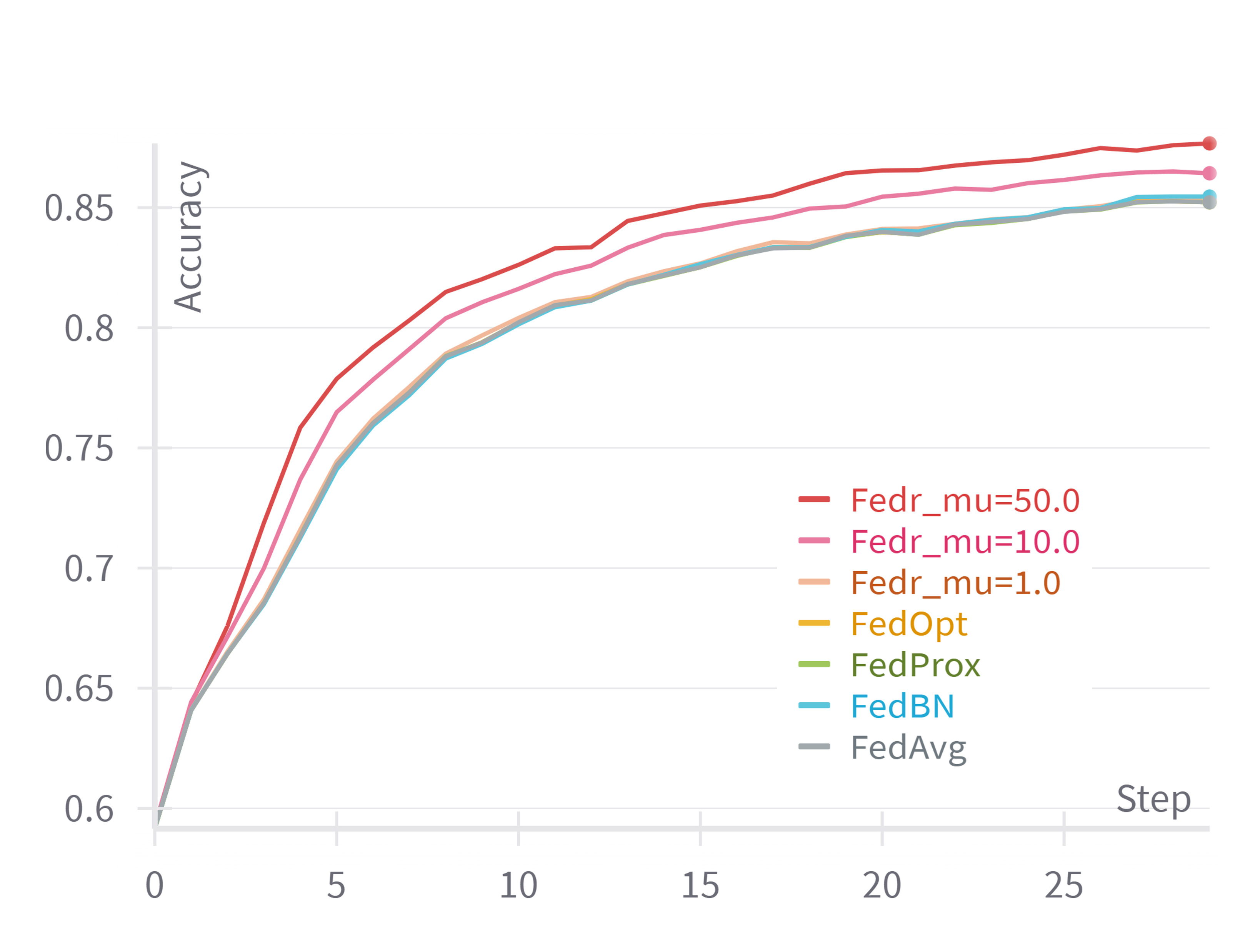}
\label{fig:exp_femnist_1_test_acc}
}\hspace{-2mm}
\subfigure[FEMNIST test\_avg\_loss]{
\includegraphics[width=0.48\textwidth]{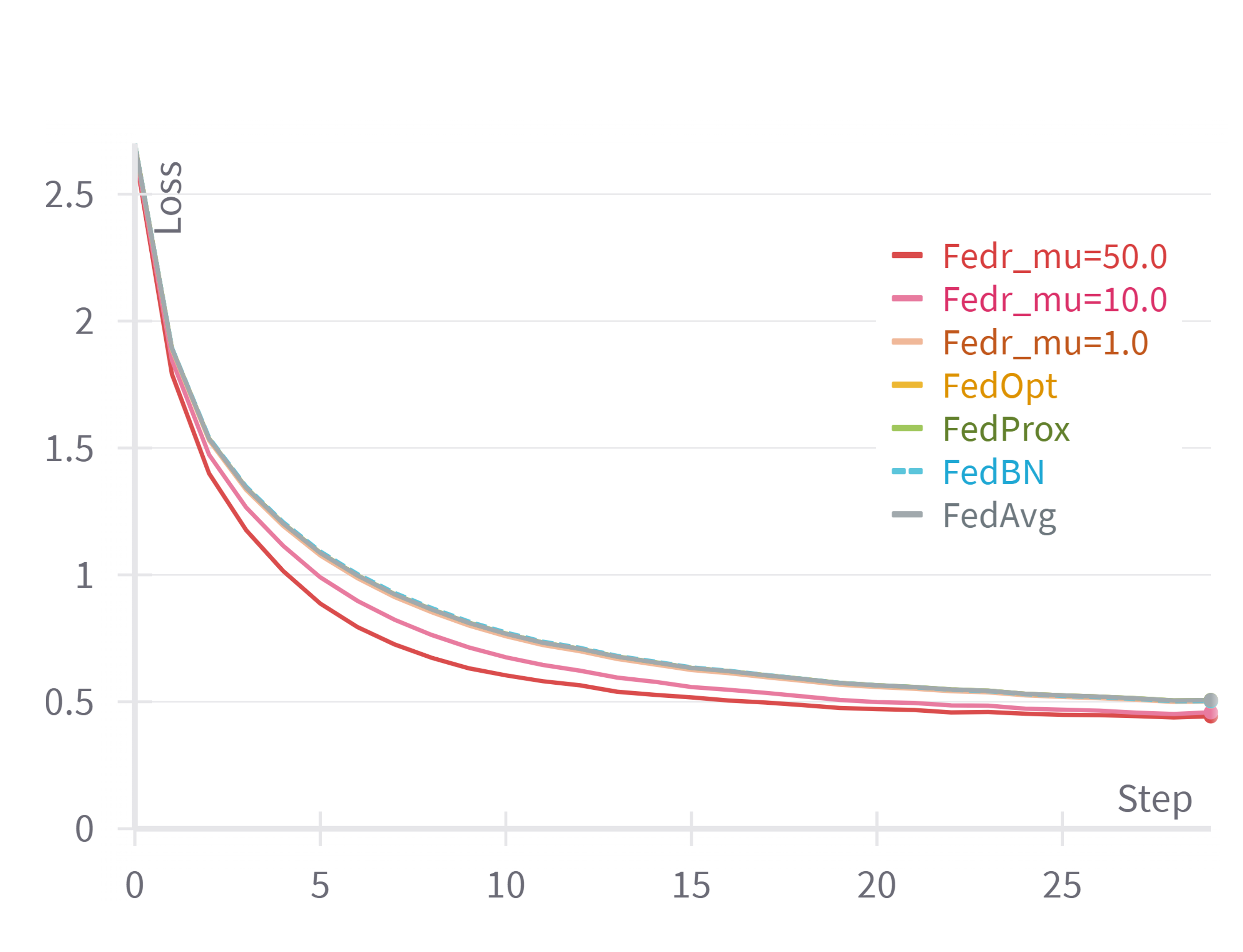}
\label{fig:exp_femnist_1_test_avg_loss}
}\hspace{-2mm}
\subfigure[CIFAR-10 test\_acc 1]{
\includegraphics[width=0.48\textwidth]{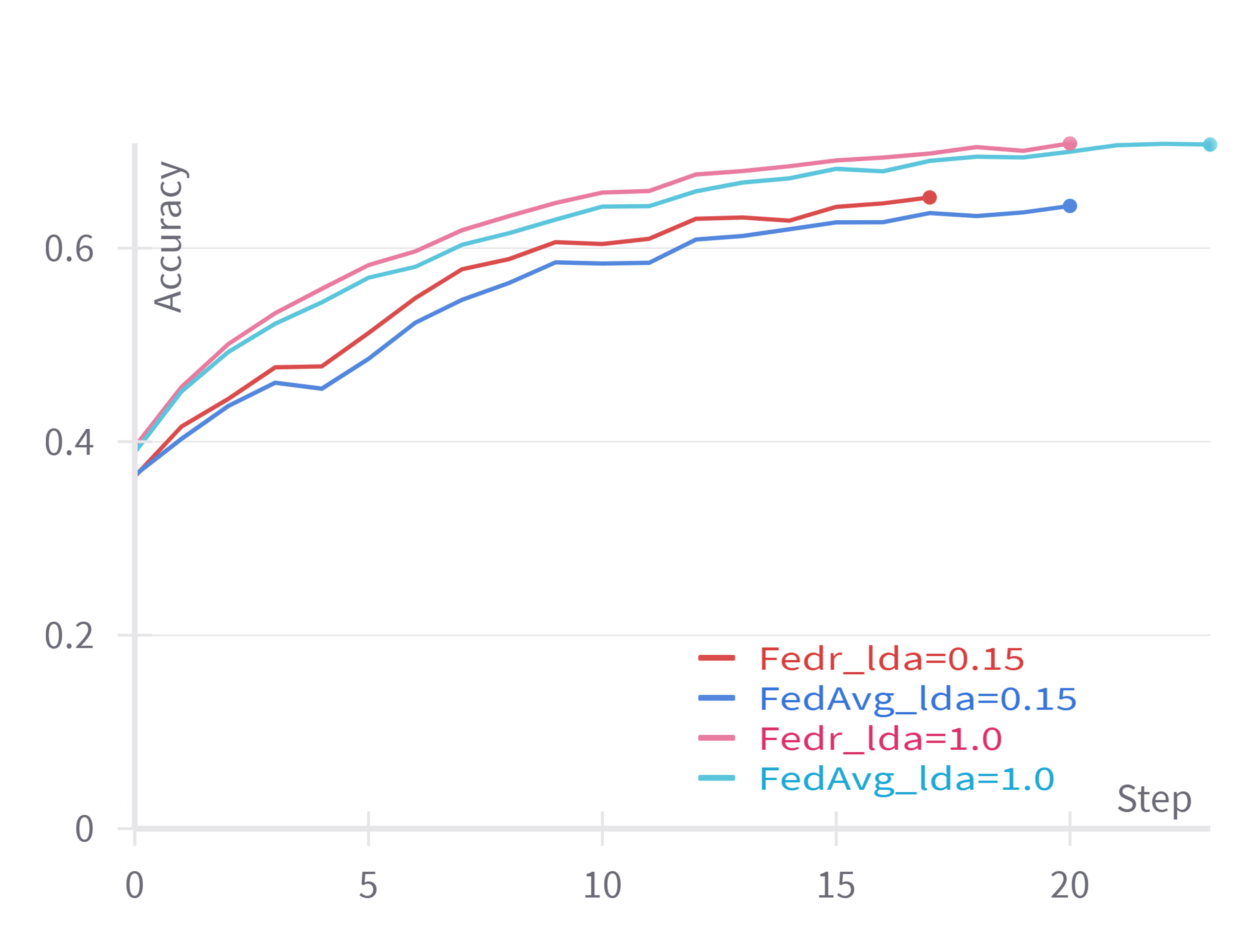}
\label{fig:exp_cifar10_1_test_acc}
}\hspace{-2mm}
\subfigure[CIFAR-10 test\_avg\_loss 1]{
\includegraphics[width=0.48\textwidth]{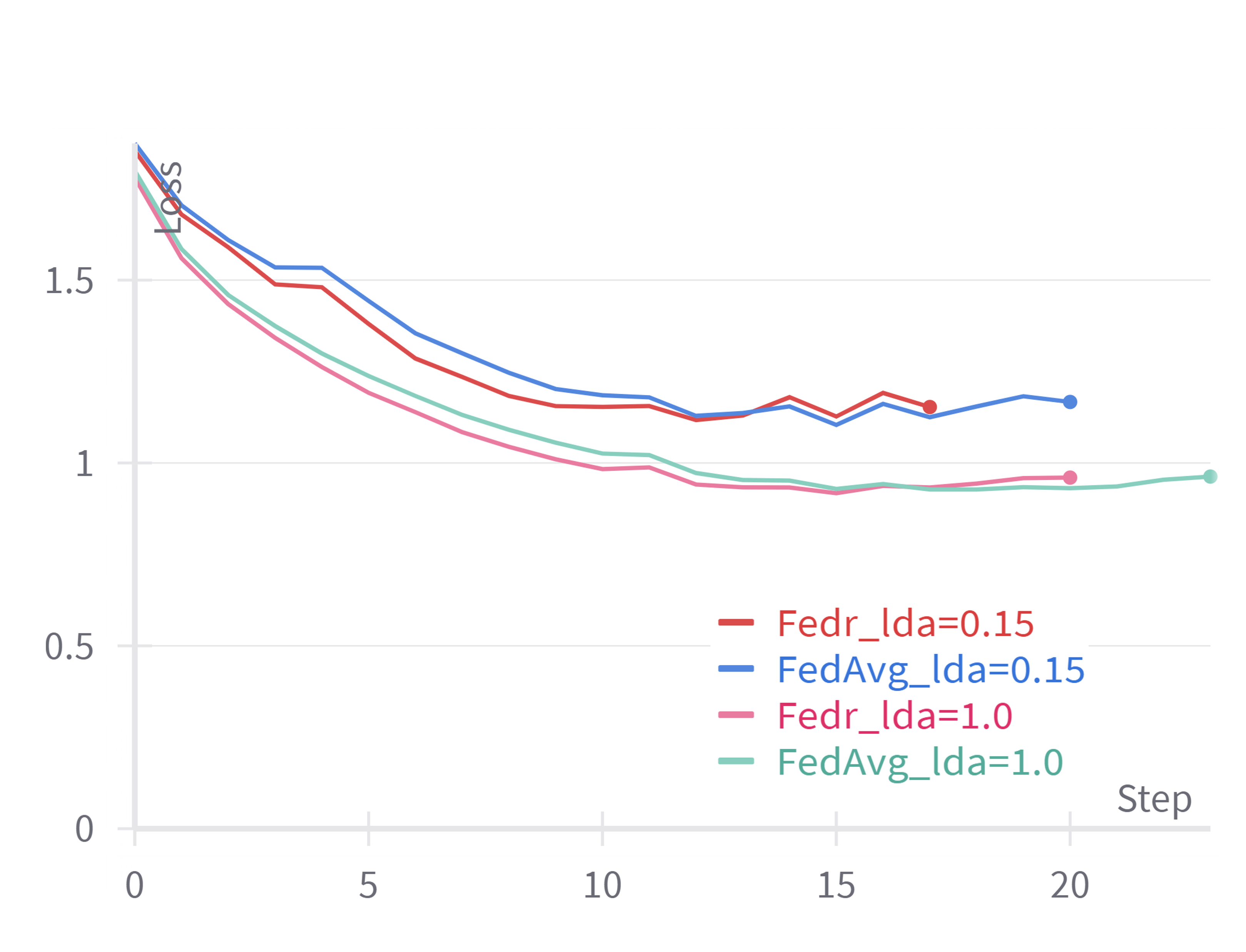}
\label{fig:exp_cifar10_1_test_avg_loss}
}\hspace{-2mm}
\subfigure[CIFAR-10 test\_acc 2]{
\includegraphics[width=0.48\textwidth]{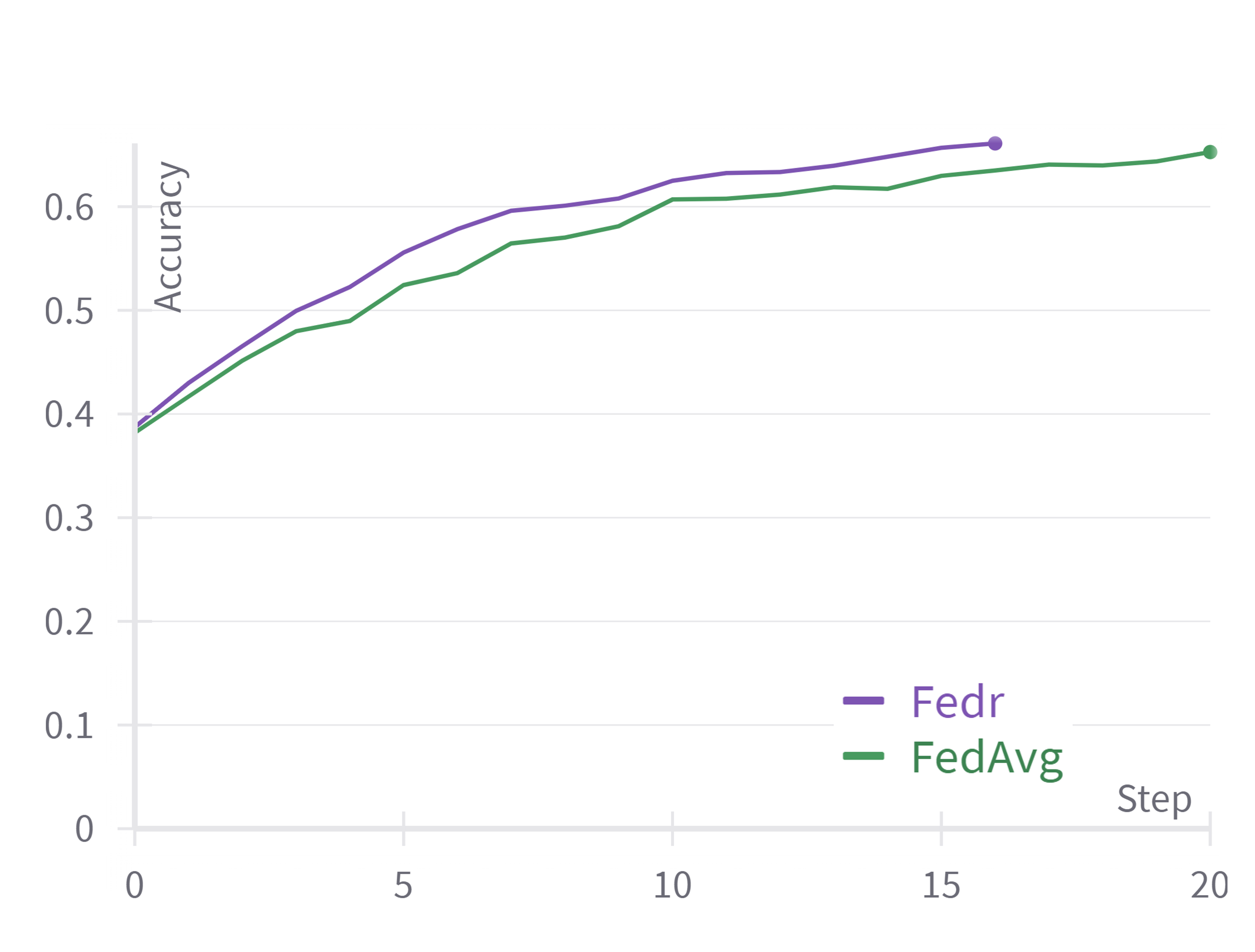}
\label{fig:exp_cifar10_2_test_acc}
}\hspace{-2mm}
\subfigure[CIFAR-10 test\_avg\_loss 2]{
\includegraphics[width=0.48\textwidth]{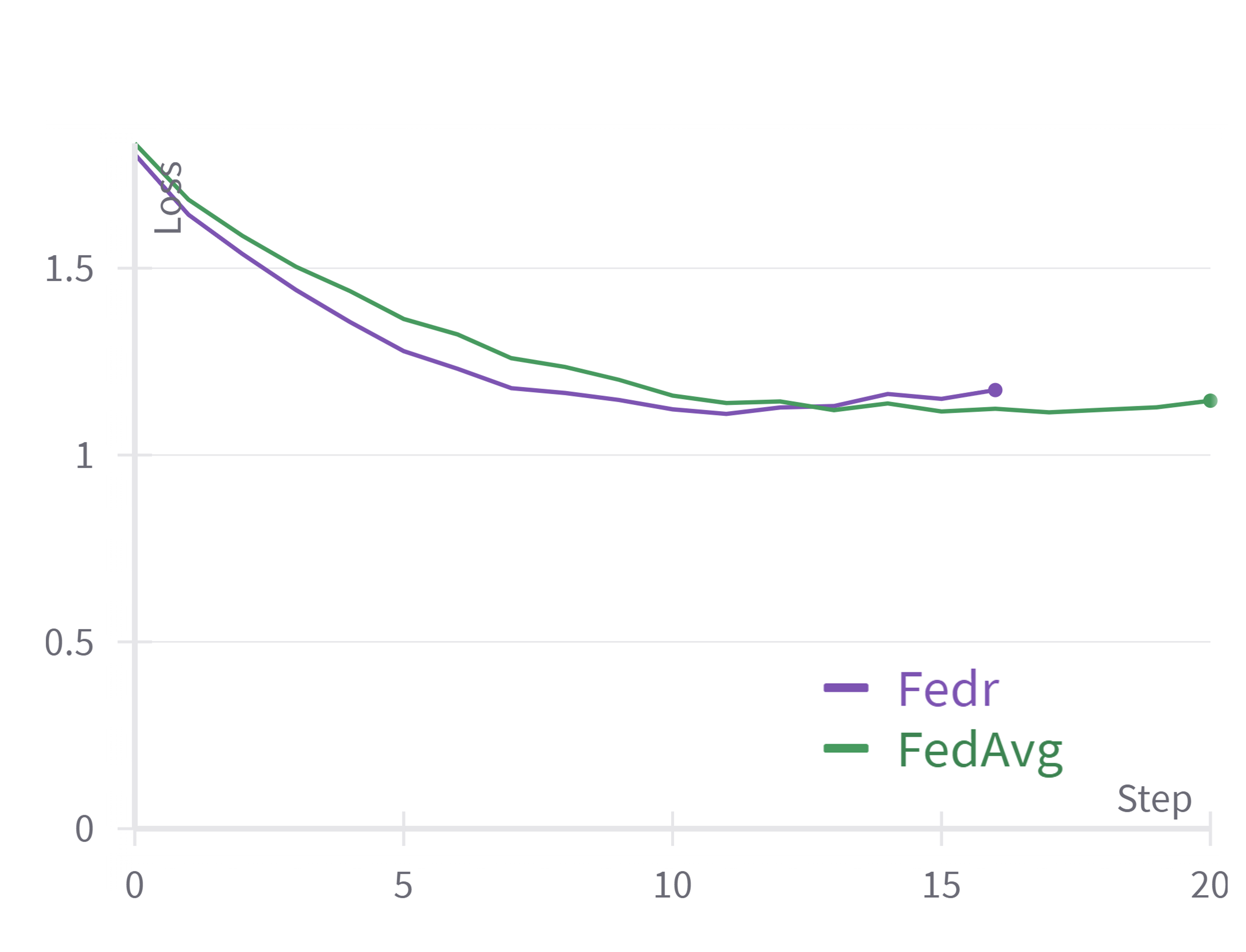}
\label{fig:exp_cifar10_2_test_avg_loss}
}
\caption{Results of Fedr on FEMNIST and CIFAR-10.}
\end{figure}

We conducted evaluations in seven different settings. In setting 1, we compared our delta regularization method with several baseline methods, including FedAvg \cite{McMahan2016CommunicationEfficientLO}, FedOpt \cite{Reddi2020AdaptiveFO}, FedBN \cite{Li2021FedBNFL}, and FedProx \cite{Sahu2018FederatedOI}, on the FEMNIST dataset with varying values of $\mu$. For FedBN \cite{Li2021FedBNFL}, we augmented the local training layers with "norms", while for FedProx \cite{Sahu2018FederatedOI}, we set $\mu$ to 0.01. The results are presented in Figures \ref{fig:exp_femnist_1_test_acc} and \ref{fig:exp_femnist_1_test_avg_loss}. The findings from setting 1 suggest that these baseline methods perform poorly when the data is non-IID. In contrast, our model delta regularization method demonstrates superior performance with lower loss. However, it's worth noting that the hyperparameter $\mu$ significantly influences the method's performance, indicating that our model delta regularization method requires meticulous manual tuning.

In setting 2, we compared our model delta regularization method with FedAvg \cite{McMahan2016CommunicationEfficientLO} on the CIFAR-10 \cite{krizhevsky2009learning} dataset. We set the $\alpha$ of the LDA to 1.0 and 0.15 for IID and non-IID data distributions, respectively. For our model delta regularization method, we evaluated both Option I (with coefficients set to [0.2, 0.3, 0.5]) and Option II (with coefficient set to 0.5), while setting $\mu$ to 0.1. The results are presented in Figures \ref{fig:exp_cifar10_1_test_acc} and \ref{fig:exp_cifar10_1_test_avg_loss}.

In setting 3, we set the SCR (sampled client ratio) to 0.5, meaning that we selected 50\% of the clients in each round. The other settings remained the same as in setting 2, and the results are shown in Figures \ref{fig:exp_cifar10_2_test_acc} and \ref{fig:exp_cifar10_2_test_avg_loss}. Our model delta regularization method exhibits a slight performance improvement over FedAvg, particularly noticeable when dealing with non-IID data in both settings. Furthermore, as the number of clients increases, our method demonstrates greater advantages. Additionally, our method shows faster convergence speeds with both IID and non-IID data.

\begin{figure}[htbp]
\centering
\subfigure[test\_avg\_imp\_ratio 1]{
\includegraphics[width=0.48\textwidth]{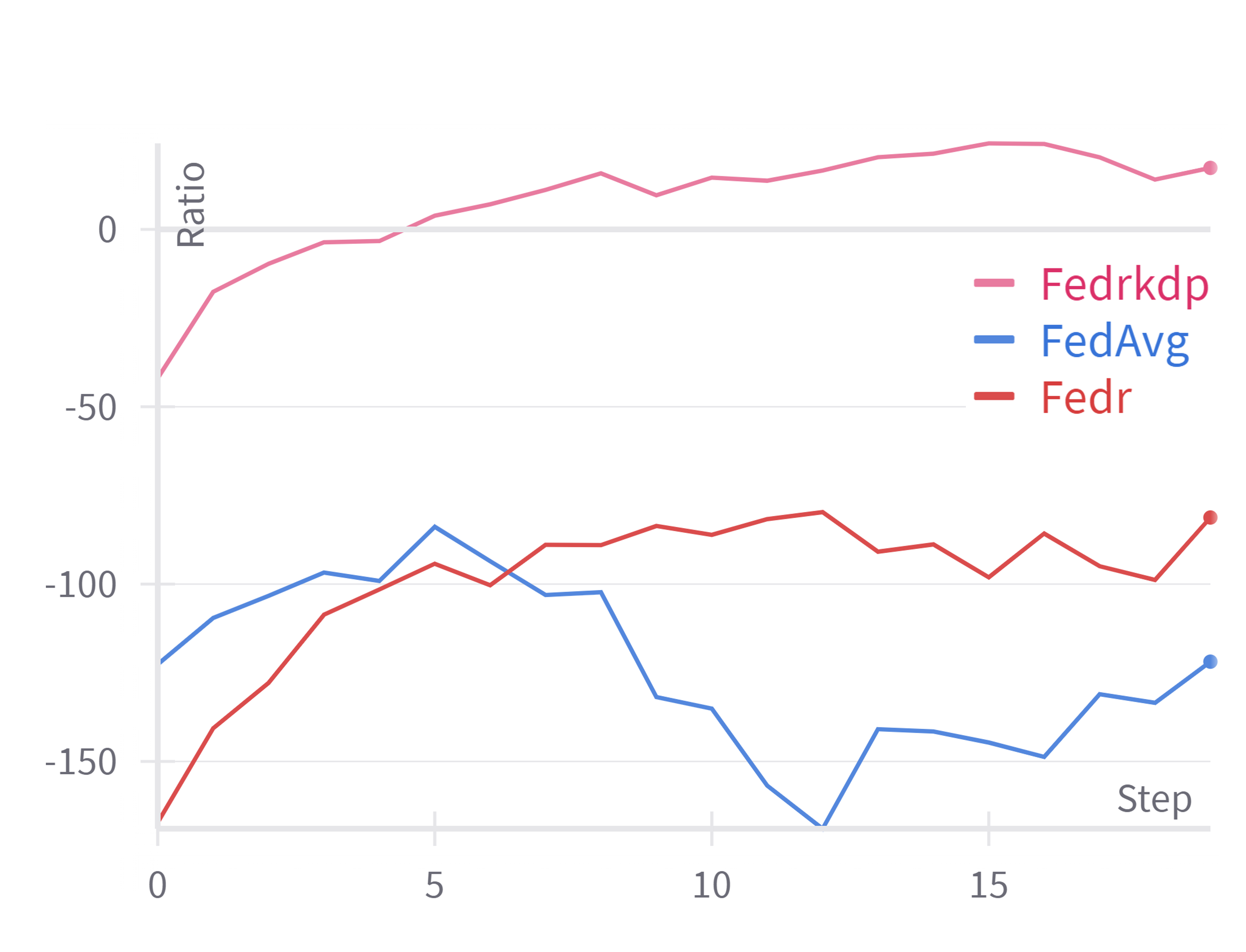}
\label{fig:exp_cikmcup_1_test_imp_ratio}
}\hspace{-2mm}
\subfigure[test\_avg\_loss 1]{
\includegraphics[width=0.48\textwidth]{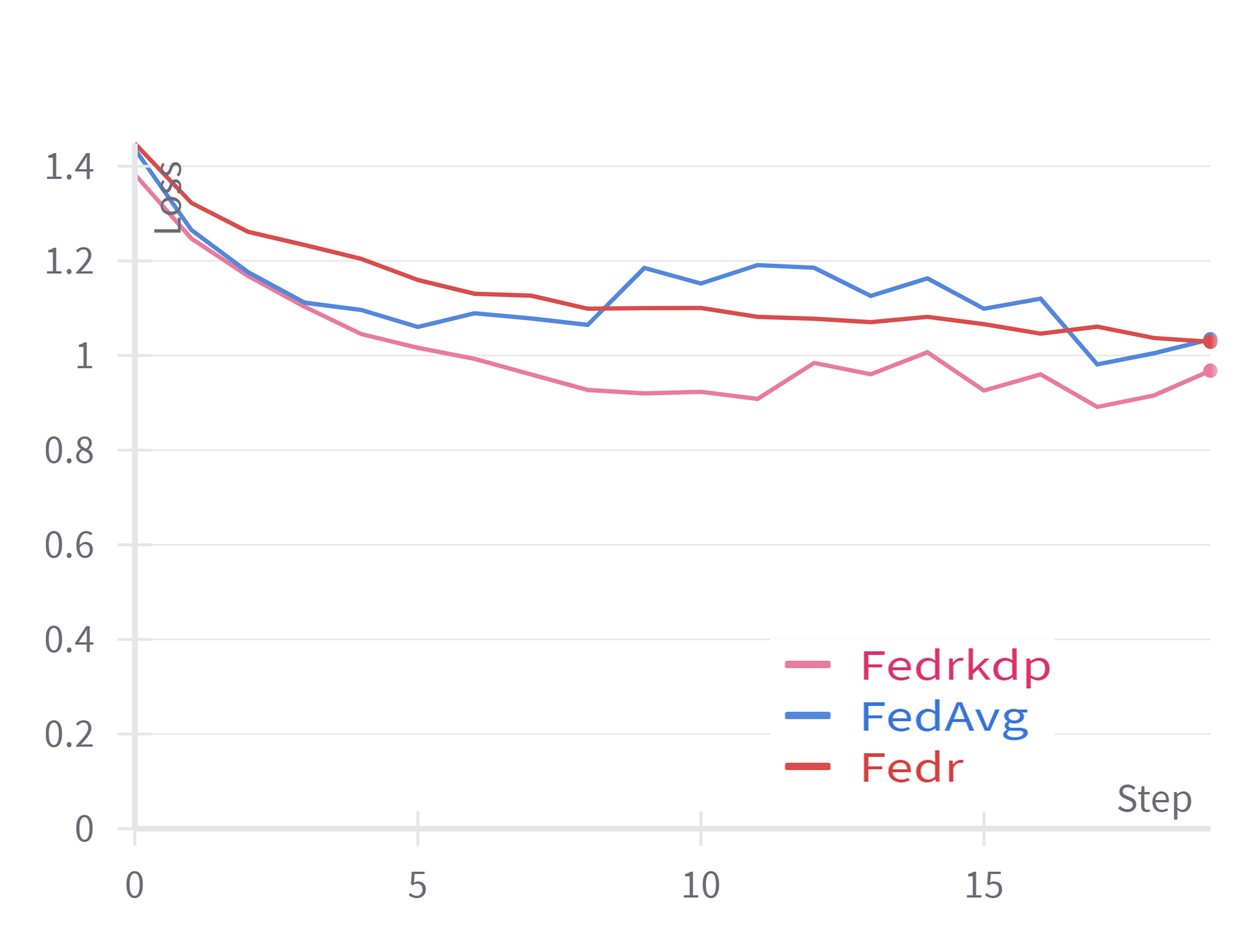}
\label{fig:exp_cikmcup_1_test_avg_loss}
}\hspace{-2mm}
\subfigure[test\_avg\_imp\_ratio 2]{
\includegraphics[width=0.48\textwidth]{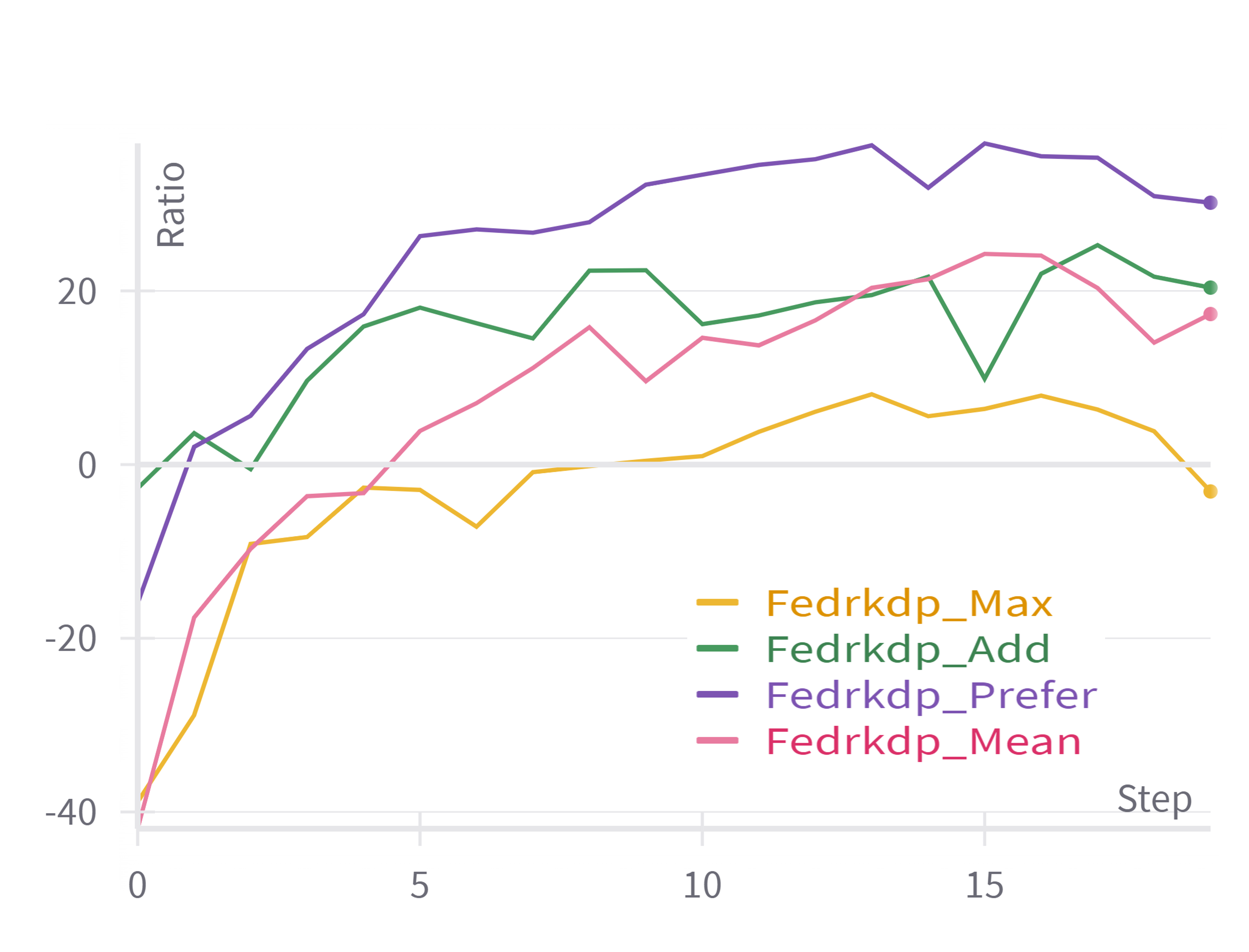}
\label{fig:exp_cikmcup_2_test_imp_ratio}
}\hspace{-2mm}
\subfigure[test\_avg\_loss 2]{
\includegraphics[width=0.48\textwidth]{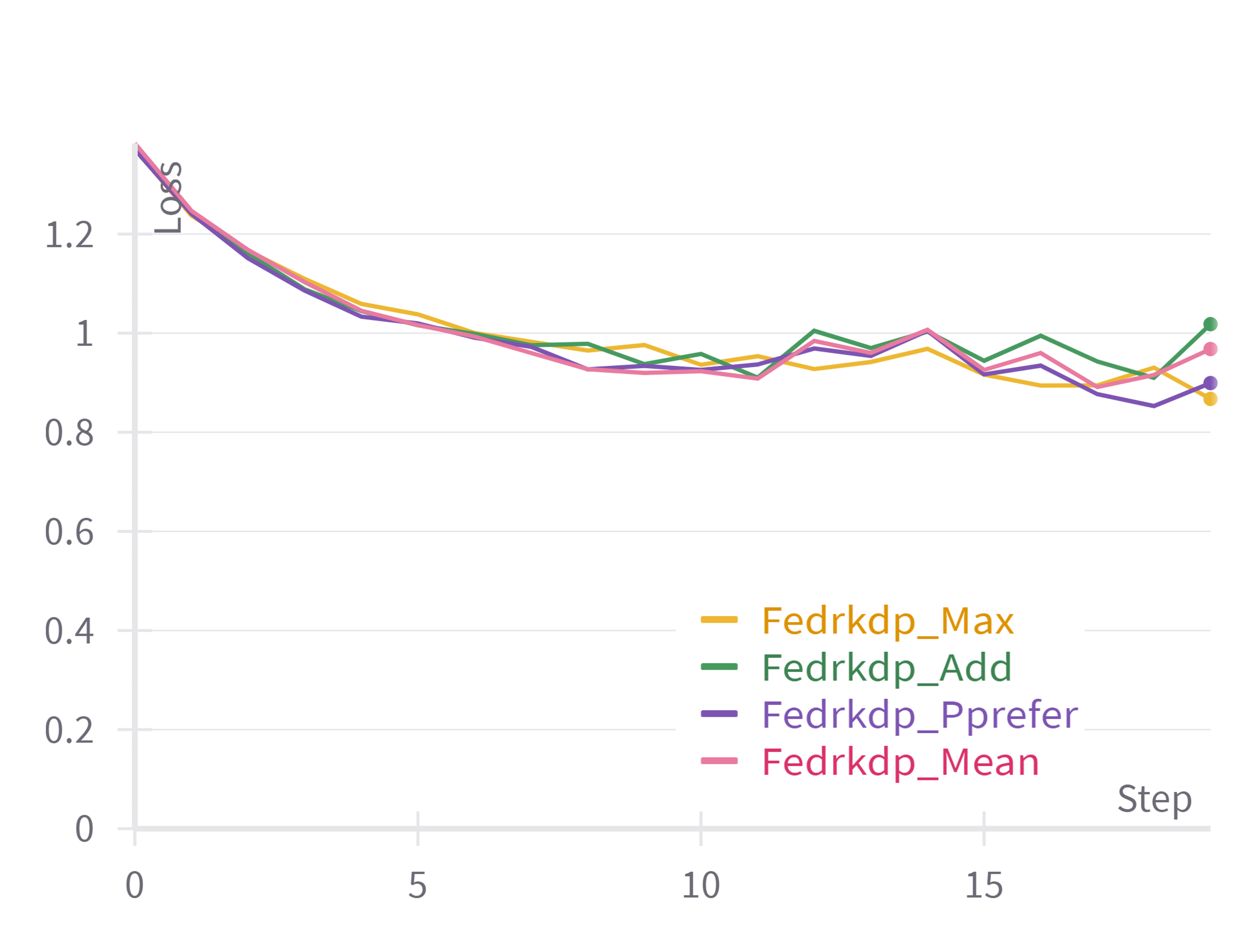}
\label{fig:exp_cikmcup_2_test_avg_loss}
}\hspace{-2mm}
\subfigure[test\_avg\_imp\_ratio 3]{
\includegraphics[width=0.48\textwidth]{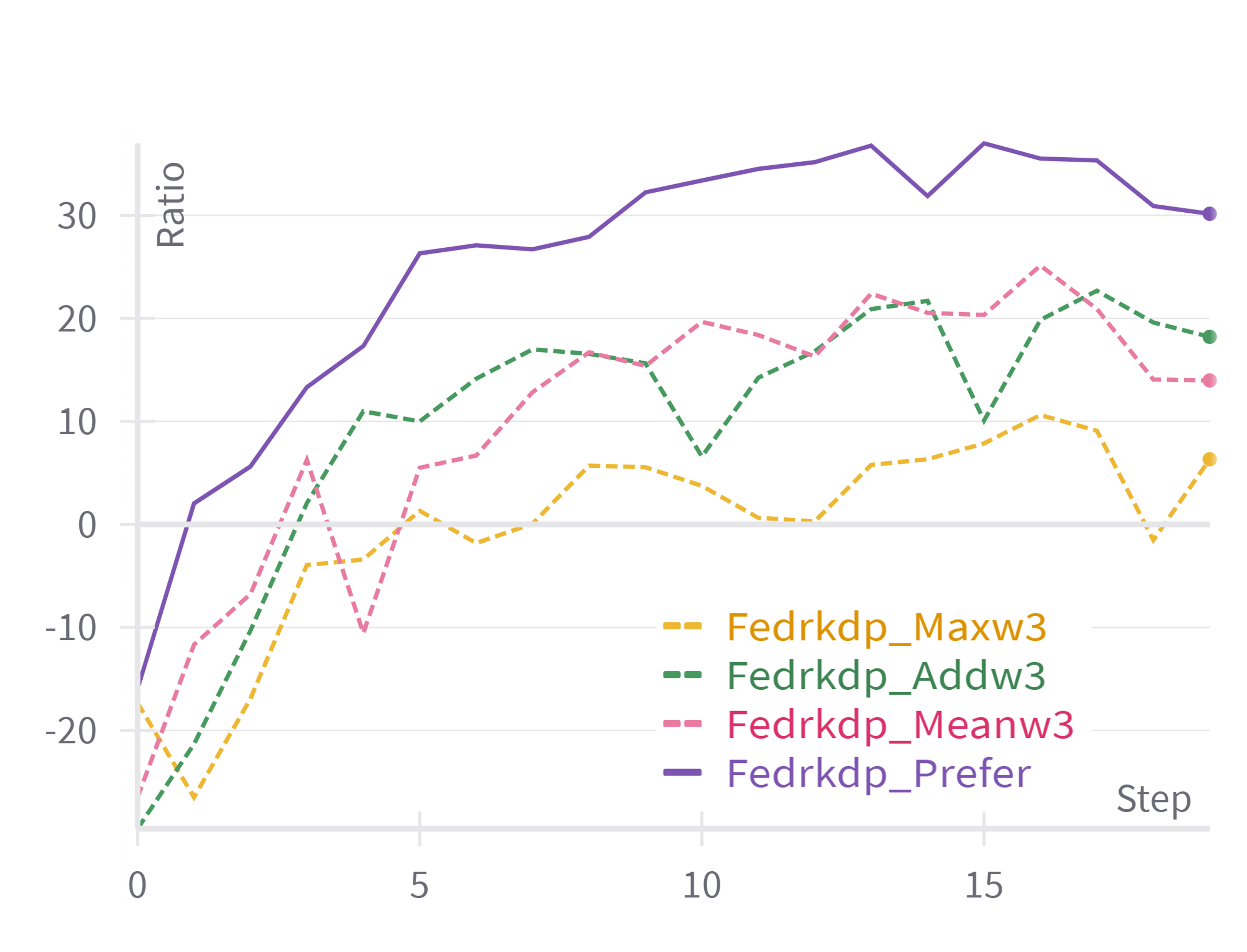}
\label{fig:exp_cikmcup_3_test_imp_ratio}
}\hspace{-2mm}
\subfigure[test\_avg\_loss 3]{
\includegraphics[width=0.48\textwidth]{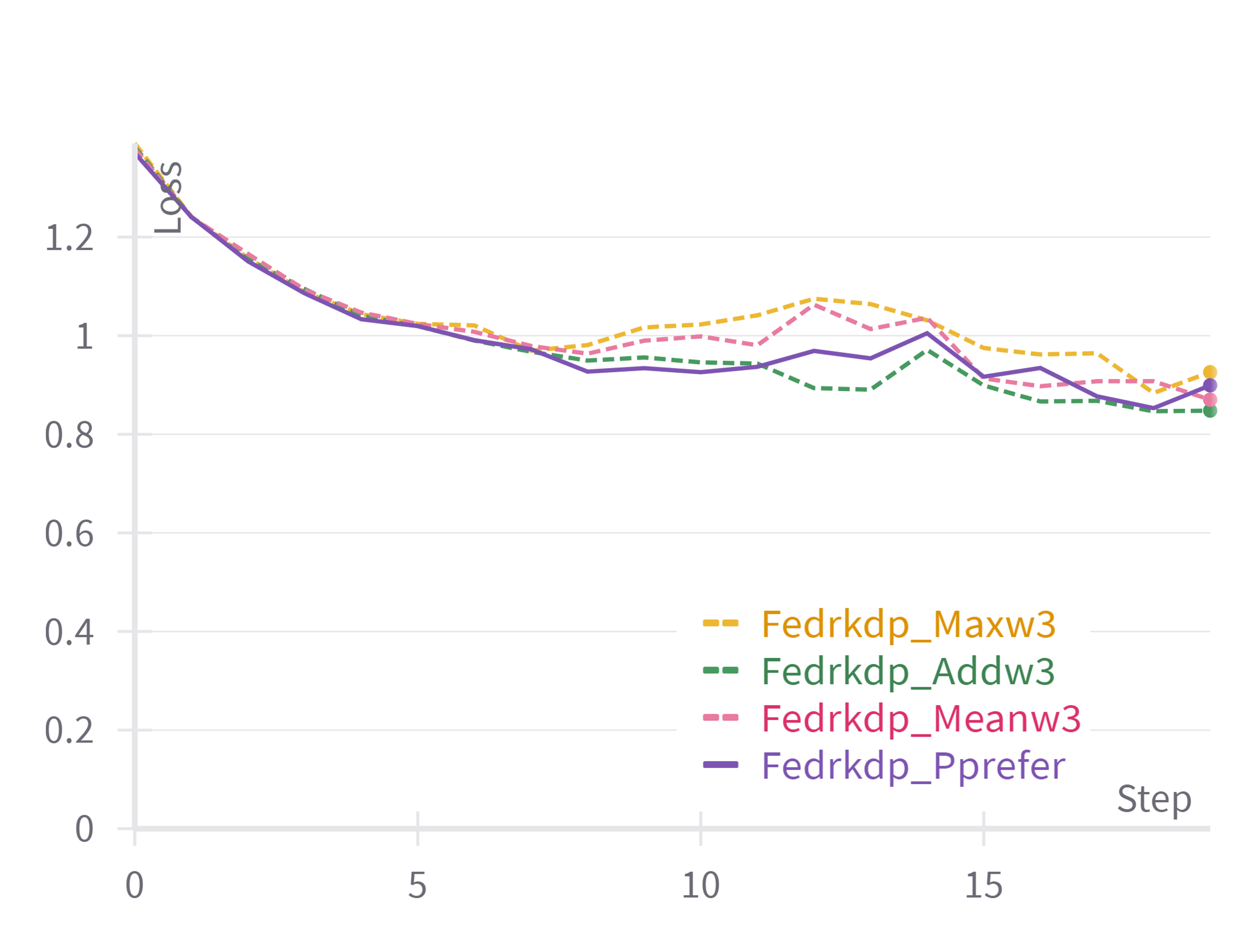}
\label{fig:exp_cikmcup_3_test_avg_loss}
}\hspace{-2mm}
\caption{Results of Fedrkdp on the CIKM22Cup.}
\label{fig:cikmcup_results}
\end{figure}

In setting 4, we evaluated the model delta regularization method, the federated knowledge distillation method, and FedAvg \cite{McMahan2016CommunicationEfficientLO} on the CIKM22Cup dataset \cite{CIKM22}. We chose Option II with the coefficient set to 0.5, and $\mu$ was chosen to be client-dependent. Additionally, $\alpha$ was set to 0.5, and $\tau$ was set to 10.0. We utilized the shared model as the personalization model and "mean" pooling as the readout operations. The results are depicted in Figures \ref{fig:exp_cikmcup_1_test_imp_ratio} and \ref{fig:exp_cikmcup_1_test_avg_loss}. Since the CIKM22Cup dataset \cite{CIKM22} simulates more extreme heterogeneity, FedAvg \cite{McMahan2016CommunicationEfficientLO} struggles to converge. In contrast, our model delta regularization provides more stable aggregations, and the federated knowledge distillation method significantly enhances performance.

\begin{table}[t]
\caption{Clients' Preference for Pooling Methods}
\label{tab:pooling_preference}
\renewcommand\tabcolsep{15pt} 
\centering
\begin{tabular}{ccccc}
\hline
Client ID & sum & mean & max & mix-pooling\\\hline
1 & 0.2818 & 0.6475 & \textbf{0.1264} & 0.1724\\
2 & 0.5822 & \textbf{0.5309} & 0.6600 & 0.5751\\
3 & 0.5280 & 0.5575 & 0.4908 & \textbf{0.4848}\\
4 & 0.3887 & \textbf{0.3075} & 0.3501 & 0.3153\\
5 & 0.6884 & \textbf{0.6332} & 0.6513 & 0.6751\\
6 & 0.4234 & 0.4369 & 0.4214 & \textbf{0.4042}\\
7 & 0.5345 & 0.5305 & \textbf{0.5156} & 0.5799\\
8 & 1.0919 & \textbf{0.4035} & 0.7545 & 0.4461\\
9 & 0.0490 & 0.0337 & 0.1211 & \textbf{0.0325}\\
10 & 0.0043 & 0.0057 & 0.0050 & \textbf{0.0036}\\
11 & \textbf{0.6445} & 0.6999 & 0.8369 & 0.9813\\
12 & 0.7730 & 1.1413 & 1.7948 & \textbf{0.5010}\\
13 & 0.0014 & 0.0029 & 0.0020 & \textbf{0.0010}\\
\hline
\end{tabular}
\end{table}

In setting 5, we assessed our mix-pooling method on the CIKM22Cup dataset \cite{CIKM22}. Clients were trained locally for 100 rounds, with "mean," "max," and mix-pooling utilized as their respective graph readout operations instead of aggregation operations. The results are summarized in Table \ref{tab:pooling_preference}. The findings show that each client exhibits a preference for a specific readout operation. Clients 9, 10, 12, and 13, which are tasked with regression, favor the mix-pooling readout operation. However, it is noteworthy that 50\% of the clients still prefer the traditional readout operations. This underscores the importance of selecting the appropriate readout operation tailored to the specific task and dataset to optimize performance.

In setting 6, we evaluated various readout operations under the federated knowledge distillation method. The personalization models utilized "mean," "sum," "max," and the preferences obtained in setting 5 as their respective readout operations. Other settings remained consistent with setting 4. The results are depicted in Figures \ref{fig:exp_cikmcup_2_test_imp_ratio} and \ref{fig:exp_cikmcup_2_test_avg_loss}. The federated knowledge distillation method demonstrates the capability to enhance performance with any readout operation. Notably, we observed that the best performance is achieved when each client utilizes its preferred readout operation.

Due to the increased width of the linear layers in mix-pooling compared to regular readout operations, we conducted an additional evaluation experiment in setting 7 to assess the effect of layer width. We compared preference-based readout operation selection, mix-pooling, and "mean\_3w", "sum\_3w", and "max\_3w" layers, each with three times the width of the linear layers. Other settings remained consistent with setting 6. Additional detailed results are presented in Figures \ref{fig:exp_cikmcup_3_test_imp_ratio} and \ref{fig:exp_cikmcup_3_test_avg_loss}.

Overall, we compared our proposed methods with baseline approaches on diverse datasets. Our model delta regularization method demonstrated superior performance over baseline methods on non-IID data. Federated knowledge distillation consistently improved performance across various readout operations. Additionally, preference-based readout operation selection outperformed other methods, highlighting the importance of personalized approaches in FL.

\section{Conclusion}
This study introduced a novel federated learning optimization algorithm aimed at addressing heterogeneity within FL settings. The algorithm integrates model delta regularization, personalized models, federated knowledge distillation, and mix-pooling. Model delta regularization facilitates server-side computation of model deltas, enabling updates to clients aligned with the server's direction at low communication costs. Personalized models and federated knowledge distillation are utilized to tackle task heterogeneity, while mix-pooling accommodates varying sensitivities of readout operations. Experimental results confirmed the effectiveness of these methods. In future work, we plan to explore FL applications within the healthcare domain. 

\section*{Acknowledgements}
This research was partially supported by KAKENHI (20K11830) Japan.

\bibliographystyle{unsrt}
\bibliography{reference}

\begin{thebibliography}{10}

\bibitem{li2023spotgan}
Chen Li and Yoshihiro Yamanishi.
\newblock {SpotGAN}: A reverse-transformer gan generates scaffold-constrained molecules with property optimization.
\newblock In {\em Joint European Conference on Machine Learning and Knowledge Discovery in Databases}, pages 323--338. Springer, 2023.

\bibitem{li2024tengan}
Chen Li and Yoshihiro Yamanishi.
\newblock {TenGAN}: Pure transformer encoders make an efficient discrete gan for de novo molecular generation.
\newblock In {\em International Conference on Artificial Intelligence and Statistics}, pages 361--369. PMLR, 2024.

\bibitem{Murphy2018TheGD}
J.~F.~A. Murphy.
\newblock The general data protection regulation (gdpr).
\newblock {\em Irish medical journal}, 111(5):747, 2018.

\bibitem{McMahan2016CommunicationEfficientLO}
H.~B. McMahan, Eider Moore, Daniel Ramage, Seth Hampson, and Blaise~Ag{\"u}era y~Arcas.
\newblock Communication-efficient learning of deep networks from decentralized data.
\newblock In {\em International Conference on Artificial Intelligence and Statistics}, 2016.

\bibitem{McMahan2016FederatedLO}
H.~B. McMahan, Eider Moore, Daniel Ramage, and Blaise~Ag{\"u}era y~Arcas.
\newblock Federated learning of deep networks using model averaging.
\newblock {\em arXiv}, abs/1602.05629, 2016.

\bibitem{duan2020learning}
Rui Duan, Mary~Regina Boland, Zixuan Liu, Yue Liu, Howard~H Chang, Hua Xu, Haitao Chu, Christopher~H Schmid, Christopher~B Forrest, John~H Holmes, et~al.
\newblock Learning from electronic health records across multiple sites: A communication-efficient and privacy-preserving distributed algorithm.
\newblock {\em Journal of the American Medical Informatics Association}, 27(3):376--385, 2020.

\bibitem{huang2019patient}
Li~Huang, Andrew~L Shea, Huining Qian, Aditya Masurkar, Hao Deng, and Dianbo Liu.
\newblock Patient clustering improves efficiency of federated machine learning to predict mortality and hospital stay time using distributed electronic medical records.
\newblock {\em Journal of biomedical informatics}, 99:103291, 2019.

\bibitem{li2019distributed}
Ziyi Li, Kirk Roberts, Xiaoqian Jiang, and Qi~Long.
\newblock Distributed learning from multiple ehr databases: contextual embedding models for medical events.
\newblock {\em Journal of biomedical informatics}, 92:103138, 2019.

\bibitem{Kairouz2019AdvancesAO}
Peter~Kairouz et. al.
\newblock Advances and open problems in federated learning.
\newblock {\em Found. Trends Mach. Learn.}, 14:1--210, 2019.

\bibitem{Li2019FederatedLC}
Tian Li, Anit~Kumar Sahu, Ameet Talwalkar, and Virginia Smith.
\newblock Federated learning: Challenges, methods, and future directions.
\newblock {\em IEEE Signal Processing Magazine}, 37:50--60, 2019.

\bibitem{Gao2022ASO}
Dashan Gao, Xin Yao, and Qian Yang.
\newblock A survey on heterogeneous federated learning.
\newblock {\em arXiv}, abs/2210.04505, 2022.

\bibitem{Zhao2018FederatedLW}
Yue Zhao, Meng Li, Liangzhen Lai, Naveen Suda, Damon Civin, and Vikas Chandra.
\newblock Federated learning with non-iid data.
\newblock {\em arXiv}, abs/1806.00582, 2018.

\bibitem{Hsu2019MeasuringTE}
Tzu-Ming~Harry Hsu, Qi, and Matthew Brown.
\newblock Measuring the effects of non-identical data distribution for federated visual classification.
\newblock {\em arXiv}, abs/1909.06335, 2019.

\bibitem{Sahu2018FederatedOI}
Anit~Kumar Sahu, Tian Li, Maziar Sanjabi, Manzil Zaheer, Ameet Talwalkar, and Virginia Smith.
\newblock Federated optimization in heterogeneous networks.
\newblock {\em arXiv: Learning}, 2018.

\bibitem{Zhang2020FedPDAF}
Xinwei Zhang, Mingyi Hong, Sairaj~V. Dhople, Wotao Yin, and Yang Liu.
\newblock Fedpd: A federated learning framework with optimal rates and adaptivity to non-iid data.
\newblock {\em arXiv}, abs/2005.11418, 2020.

\bibitem{Li2019FedDANEAF}
Tian Li, Anit~Kumar Sahu, Manzil Zaheer, Maziar Sanjabi, Ameet Talwalkar, and Virginia Smith.
\newblock Feddane: A federated newton-type method.
\newblock {\em 2019 53rd Asilomar Conference on Signals, Systems, and Computers}, pages 1227--1231, 2019.

\bibitem{Karimireddy2019SCAFFOLDSC}
Sai~Praneeth Karimireddy, Satyen Kale, Mehryar Mohri, Sashank~J. Reddi, Sebastian~U. Stich, and Ananda~Theertha Suresh.
\newblock Scaffold: Stochastic controlled averaging for federated learning.
\newblock In {\em International Conference on Machine Learning}, 2019.

\bibitem{Li2021FedBNFL}
Xiaoxiao Li, Meirui Jiang, Xiaofei Zhang, Michael Kamp, and Qi~Dou.
\newblock Fedbn: Federated learning on non-iid features via local batch normalization.
\newblock {\em arXiv}, abs/2102.07623, 2021.

\bibitem{Liang2020ThinkLA}
Paul~Pu Liang, Terrance Liu, Liu Ziyin, Ruslan Salakhutdinov, and Louis-Philippe Morency.
\newblock Think locally, act globally: Federated learning with local and global representations.
\newblock {\em arXiv}, abs/2001.01523, 2020.

\bibitem{Shen2022CD2pFedCD}
Yiqing Shen, Yuyin Zhou, and Lequan Yu.
\newblock Cd2-pfed: Cyclic distillation-guided channel decoupling for model personalization in federated learning.
\newblock {\em 2022 IEEE/CVF Conference on Computer Vision and Pattern Recognition (CVPR)}, pages 10031--10040, 2022.

\bibitem{Li2020DittoFA}
Tian Li, Shengyuan Hu, Ahmad Beirami, and Virginia Smith.
\newblock Ditto: Fair and robust federated learning through personalization.
\newblock In {\em International Conference on Machine Learning}, 2020.

\bibitem{Reddi2020AdaptiveFO}
Sashank~J. Reddi, Zachary~B. Charles, Manzil Zaheer, Zachary Garrett, Keith Rush, Jakub Konecn{\'y}, Sanjiv Kumar, and H.~B. McMahan.
\newblock Adaptive federated optimization.
\newblock {\em arXiv}, abs/2003.00295, 2020.

\bibitem{Acar2021FederatedLB}
Durmus Alp~Emre Acar, Yue Zhao, Ramon~Matas Navarro, Matthew Mattina, Paul~N. Whatmough, and Venkatesh Saligrama.
\newblock Federated learning based on dynamic regularization.
\newblock {\em arXiv}, abs/2111.04263, 2021.

\bibitem{Collins2021ExploitingSR}
Liam Collins, Hamed Hassani, Aryan Mokhtari, and Sanjay Shakkottai.
\newblock Exploiting shared representations for personalized federated learning.
\newblock In {\em International Conference on Machine Learning}, 2021.

\bibitem{Li2019FedMDHF}
Daliang Li and Junpu Wang.
\newblock Fedmd: Heterogenous federated learning via model distillation.
\newblock {\em arXiv}, abs/1910.03581, 2019.

\bibitem{Yao2021LocalGlobalKD}
Dezhong Yao, Wanning Pan, Yutong Dai, Yao Wan, Xiaofeng Ding, Hai Jin, Zheng Xu, and Lichao Sun.
\newblock Local-global knowledge distillation in heterogeneous federated learning with non-iid data.
\newblock 2021.

\bibitem{Bucila2006ModelC}
Cristian Bucila, Rich Caruana, and Alexandru Niculescu-Mizil.
\newblock Model compression.
\newblock In {\em Knowledge Discovery and Data Mining}, 2006.

\bibitem{Hinton2015DistillingTK}
Geoffrey~E. Hinton, Oriol Vinyals, and Jeffrey Dean.
\newblock Distilling the knowledge in a neural network.
\newblock {\em arXiv}, abs/1503.02531, 2015.

\bibitem{li2024gxvaes}
Chen Li and Yoshihiro Yamanishi.
\newblock {GxVAEs:} two joint vaes generate hit molecules from gene expression profiles.
\newblock In {\em Proceedings of the AAAI Conference on Artificial Intelligence}, volume~38, pages 13455--13463, 2024.

\bibitem{caldas2018leaf}
Sebastian Caldas, Sai Meher~Karthik Duddu, Peter Wu, Tian Li, Jakub Kone{\v{c}}n{\`y}, H~Brendan McMahan, Virginia Smith, and Ameet Talwalkar.
\newblock Leaf: A benchmark for federated settings.
\newblock {\em arXiv preprint arXiv:1812.01097}, 2018.

\bibitem{krizhevsky2009learning}
Alex Krizhevsky, Geoffrey Hinton, et~al.
\newblock Learning multiple layers of features from tiny images.
\newblock 2009.

\bibitem{CIKM22}
FederatedScope.
\newblock Cikm 2022 analyticup competition: Federated hetero-task learning.
\newblock \url{https://federatedscope.io/competition/}.

\bibitem{Xie2022FederatedScopeAC}
Yuexiang Xie, Zhen Wang, Daoyuan Chen, Dawei Gao, Liuyi Yao, Weirui Kuang, Yaliang Li, Bolin Ding, and Jingren Zhou.
\newblock Federatedscope: A comprehensive and flexible federated learning platform via message passing.
\newblock {\em arXiv}, abs/2204.05011, 2022.

\end{thebibliography}
\end{document}